\def\year{2019}\relax
\documentclass[letterpaper]{article} 
\usepackage{aaai19}  
\usepackage{times}  
\usepackage{helvet}  
\usepackage{courier}  
\usepackage{url}  
\usepackage{graphicx}  
\usepackage{enumerate}
\usepackage{amsfonts}
\usepackage{amsmath}
\usepackage{amssymb}
\usepackage{subfigure}
\usepackage{fancyhdr}
\usepackage{enumerate}
\usepackage{float}
\begin{document}
%
\title{From Zero-Shot Learning to Cold-Start Recommendation}
\author{Jingjing Li$^1$, Mengmeng Jing$^1$, Ke Lu$^1$, Lei Zhu$^2$, Yang Yang$^1$, Zi Huang$^3$\\
$~^1$ University of Electronic Science and Technology of China\\
$~^2$ Shandong Normal University; $~^3$The University of Queensland\\
lijin117@yeah.net; jingmeng1992@gmail.com; kel@uestc.edu.cn; leizhu0608@gmail.com\\
}
\maketitle
\begin{abstract}
Zero-shot learning (ZSL) and cold-start recommendation (CSR) are two challenging problems in computer vision and recommender system, respectively. In general, they are independently investigated in different communities. This paper, however, reveals that ZSL and CSR are two extensions of the same intension. Both of them, for instance, attempt to predict unseen classes and involve two spaces, one for direct feature representation and the other for supplementary description. Yet there is no existing approach which addresses CSR from the ZSL perspective. This work, for the first time, formulates CSR as a ZSL problem, and a tailor-made ZSL method is proposed to handle CSR. Specifically, we propose a Low-rank Linear Auto-Encoder (LLAE), which challenges three cruxes, i.e., domain shift, spurious correlations and computing efficiency, in this paper. LLAE consists of two parts, a low-rank encoder maps user behavior into user attributes and a symmetric decoder reconstructs user behavior from user attributes. Extensive experiments on both ZSL and CSR tasks verify that the proposed method is a win-win formulation, i.e., not only can CSR be handled by ZSL models with a significant performance improvement compared with several conventional state-of-the-art methods, but the consideration of CSR can benefit ZSL as well. \footnote{Codes are available at https://github.com/lijin118/LLAE.}
\end{abstract}

\section{Introduction}
From a near-infinity inventory, recommender systems~\cite{bobadilla2013recommender,li2017two} pick a fraction of items which a user might enjoy based on the user's current context and past behavior~\cite{smith2017two}. If the past behavior, however, is not available, e.g., for a new user, most recommender systems, especially those popular ones based on collaborative filtering (CF)~\cite{ekstrand2011collaborative}, would be stuck. Different solutions have been proposed to handle this problem, which is widely known as cold-start recommendation (CSR)~\cite{lin2013addressing}. Recently, cross-domain information~\cite{fernandez2012cross}, personal information~\cite{fernandez2016alleviating} and social network data~\cite{sedhain2017low} have been used to facilitate CSR.

\begin{figure}[t]
\begin{center}
    \includegraphics[width=0.7\linewidth]{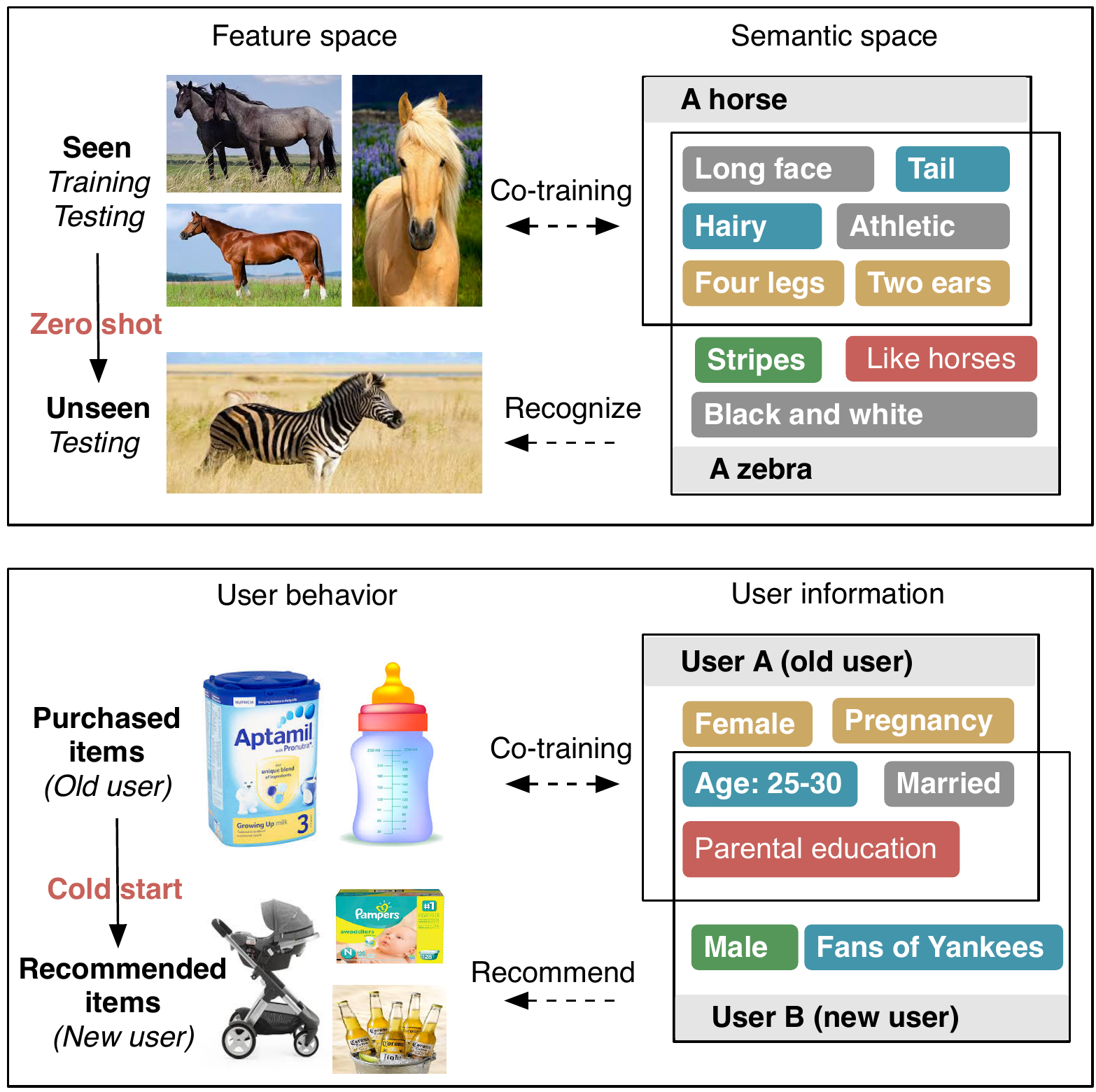}
\end{center}
\vspace{-10pt}
\caption{Illustration of ZSL and CSR. It can be seen that they are two extensions of the same intension. However, there is no existing work which addresses CSR from the ZSL perspective. For the first time, CSR is formulated as a ZSL problem and a tailor-made method is proposed to solve CSR in this paper.}
\label{fig:main}
\end{figure}

If we take a close look at previous work, it is not hard to find out that the very basic idea behind existing CSR methods~\cite{lin2013addressing,li2017two} is to {\it leverage user preferences to generate recommendations for new users}. This is quite reasonable, simply because you cannot deliver the {\it right} thing to a person you barely know. With the user preferences, we will then have two spaces, an attribute (e.g., user preferences and personal information) space and a behavior (e.g., purchase behavior and past interactions) space, in the cold-start recommendation. The attribute space is used to describe the user preferences, and the behavior space is used to represent the user interactions in the target system. As a result, cold-start recommendation can be defined as a problem to generate recommendations for a fresh user where we have nothing about the user in the behavior space but some side information about the user in the attribute space. With the assumption that {\it people with the similar preferences would have the similar consuming behavior}, cold-start recommendation can be done by two steps: 1) mapping the behavior space to the attribute space, so that the new users can be linked with the old users; 2) reconstructing user behavior by user attributes, so that we can generate recommendations for new users. Does it ring a bell now? It is a special case of zero-shot learning (ZSL)~\cite{kodirov2017semantic,dinglow2017,yang2016zero}.

This paper, for the best of our knowledge, is the first one to investigate CSR in light of ZSL. From Fig.~\ref{fig:main}, we can clearly see that CSR and ZSL are two extensions of the same intension. Specifically, {\it both of them involve two spaces, one for direct feature representation and the other for supplementary description, and both of them attempt to predict unseen cases in the feature space by leveraging the description space shared by both seen and unseen ones}. However, CSR and ZSL are not being connected ever before. In this paper, we propose a novel ZSL method to handle the CSR problem.

By formulating CSR as a ZSL problem, we challenge three cruxes in this paper. The first one is the domain shift problem. Not only are the behavior space and the attribute space heterogeneous but also the old users and the new users are divergent in probability distribution. Thus, we have to guarantee that the user behavior can be reconstructed by the user attributes. The second crux is that user behavior in CSR, differs from ZSL in computer vision tasks, is incredibly sparse. In real-world retail giants, such as amazon.com, there are hundreds of millions of users and even more items. A specific user, however, only has a small number of interactions in the system with even less items. In consequence, the user-item matrix would be pretty sparse. The last challenge lies in efficiency. Recommender systems are on-line systems, and people hate waiting.

Technically, we propose a novel ZSL model, named as Low-rank Linear AutoEncoder (LLAE), based on the encoder-decoder paradigm~\cite{kodirov2017semantic,boureau2007unified} to handle CSR problems. LLAE consists of an encoder which maps the user behavior space into the user attribute space, and a decoder which reconstructs the user behavior by the user attribute. The reconstruction part guarantees that the user behavior can be generated from user attributes, so that the domain shifts between user behavior and user attributes can be mitigated. We formulate LLAE as a linear model for the efficiency, the computational cost of our model is irrelevant to the number of samples. As a result, it can be applied to large-scale datasets. Furthermore, a low-rank constraint is deployed to handle sparse issues. Low-rank representation~\cite{li2016low} has proven to be efficient for the problem of revealing true data from corrupted observations. We know that a behavior can be linked with numerous attributes, while these attributes should have different weights, and some of the attributes are trivial. If we consider all the attributes, it may weaken the dominant factors, introduce over-fitting and relax the generalization ability. The low-rank constraint, for its mathematical property, helps reveal the dominant factors and filter out trivial connections, or in other words, spurious correlations. It is worth noting that low-rank constraint also helps align the domain shifts from the view of domain adaptation~\cite{li2018heterogeneous,li2018transfer}. In summary, the contributions of this paper are:

\begin{enumerate}[1)]

\item We reveal that CSR and ZSL are two extensions of the same intension. CSR, for the first time, is formulated and solved as a ZSL problem. Our work builds a connection between CSR and ZSL by cross domain transfer, so that the advances in the two communities can be shared. For instance, when someone who focuses on ZSL noticed our work, he might want to look through recent publications on CSR for inspiration, and vice versa. 
\item A tailor-made ZSL model, low-rank linear autoencoder (LLAE), is presented to handle the challenging CSR tasks. LLAE takes the efficiency into account, so that it suits large-scale problem.  
\item Extensive experiments on both ZSL and CSR tasks demonstrate the effectiveness of our method. Excitingly, we find that not only can CSR be handled by ZSL models with a significant performance improvement, but the consideration of CSR can benefit ZSL as well. By linking CSR and ZSL, we wish that this work will benefit both of the communities and elicit more contributions.

\end{enumerate}

\section{Related Work}

{\bf Zero-shot learning.} A basic assumption behind conventional visual recognition algorithms is that some instances of the test class are included in the training set, so that other test instances can be recognized by learning from the training samples. For a large-scale dataset, however, collecting training samples for new and rare objects is painful. A curious mind may ask if we can recognize an unseen object with some semantic description just like human beings do. To that end, zero-shot learning~\cite{zhang2016zero,dinglow2017} has been proposed. Typically, ZSL algorithms learn a projection which maps visual space to the semantic space, or the reverse. Different models are proposed based on different projection strategies. From a macro perspective, existing ZSL methods can be grouped into three categories: 1) Learning a mapping function from the visual space to the semantic space~\cite{lampert2014attribute,dinglow2017}; 2) Learning a mapping function from the semantic space to the visual space~\cite{kodirov2015unsupervised}; 3) Learning a latent space which shared by the visual domain and the semantic domain~\cite{zhang2015zero,zhang2016zero}.

{\bf Cold-start recommendation.} Among the models which address cold-start recommendation, we focus on the ones which exploit side information, e.g., user attributes, personal information and user social network data, to facilitate the cold-start problem. Those models can be roughly grouped into three categories, e.g., the similarity based models~\cite{sedhain2014social,rohani2014effective}, matrix factorization models~\cite{krohn2012multi,noel2012new} and feature mapping models~\cite{gantner2010learning}.

Matrix factorization models typically factorize the relationship matrix into two latent representations by optimizing the following objective:
\begin{equation}
  \begin{array}{c}
  \min\limits_\mathbf{U,V} \|\mathbf{Y-UV}\|_F^2 + \Omega(\mathbf{U,V}),
  \end{array} 
\end{equation}
where $\Omega$ is the regularization used to avoid over-fitting. For cold-start problems, one can learn a shared $\bf U$ or $\bf V$ from the side-information, and then use it to predict $\bf Y$.

Feature mapping models normally learn a feature mapping between the side-information and one of the latent feature representations. The differences between the matrix factorization models and the feature mapping models is that in matrix factorization models the shared $\bf U$ is jointly learned from $\bf Y$ and the the side-information, while in feature mapping models, one needs to learn an additional feature mapping, and further learn different $\bf{U}_s$ and $\bf{U}_t$ by sharing the feature mapping. More details can be found in~\cite{sedhain2017low,gantner2010learning}.

\section{Problem Formulation}
\subsection{Notations} 
In this paper, we use bold lowercase letters to represent vectors, bold uppercase letters to represent matrices. For a matrix $\mathbf M$, its Frobenius norm is defined as ${\Vert {\bf M} \Vert}\mathrm{_F} = \sqrt {\sum_i {\delta_i (\mathbf{M})}^2}$, where ${\delta_i (\mathbf{M})}$ is the $i$-th singular value of the matrix $\mathbf M$. The trace of matrix $\mathbf M$ is represented by the expression $\mathrm{tr} \mathbf{(M)}$. For clarity, we also show the frequently used notations in Table~\ref{tab:notaions}. 

\begin{table}[t]
\centering
\scriptsize
\caption{Notations and corresponding descriptions, where $n$, $d$ and $k$ denote the number of samples and dimensionality of behavior space and attribute space, $r$ is the rank of a matrix.}
\begin{tabular}{|l|l|}
\hline
Notation&Description\\
\hline
$\mathbf{X} \in \mathbb{R}^{d*n}$ & user behavior space (CSR) / visual space (ZSL) \\
$\mathbf{S} \in \mathbb{R}^{k*n}$ & user attribute space (CSR) / semantic space (ZSL)~~~ \\
$\mathbf{W} \in \mathbb{R}^{k*d}$ & encoder  \\
$\mathbf{M, W^\top} \in \mathbb{R}^{d*k}~~~$ & decoder  \\
$\mathbf{V} \in \mathbb{R}^{k*(k-r)}$ & the singular vectors of $\mathbf{WW}^\top$ \\
$\lambda>0, \beta>0$ & penalty parameters  \\
\hline
\end{tabular}
\label{tab:notaions}
\end{table}

\subsection{Linear Low-rank Denoising Autoencoder}
Given an input data matrix {\bf X}, suppose that we can learn a mapping $\bf W$ which projects matrix {\bf X} into a latent space {\bf S}, and another mapping $\bf M$ which can reconstructs $\bf X$ from $\bf S$. As an optimization problem, our aim is to minimize the reconstruction error. Thus, we have the following objective:

\begin{equation}
\label{eq:llae1}
  \begin{array}{c}
  \min\limits_\mathbf{W, M} \mathbf{\|X-MWX\|}_F^2, ~~s.t.~\mathbf{WX=S}.
  \end{array} 
\end{equation}

\begin{figure}[t]
\begin{center}
    \includegraphics[width=0.7\linewidth]{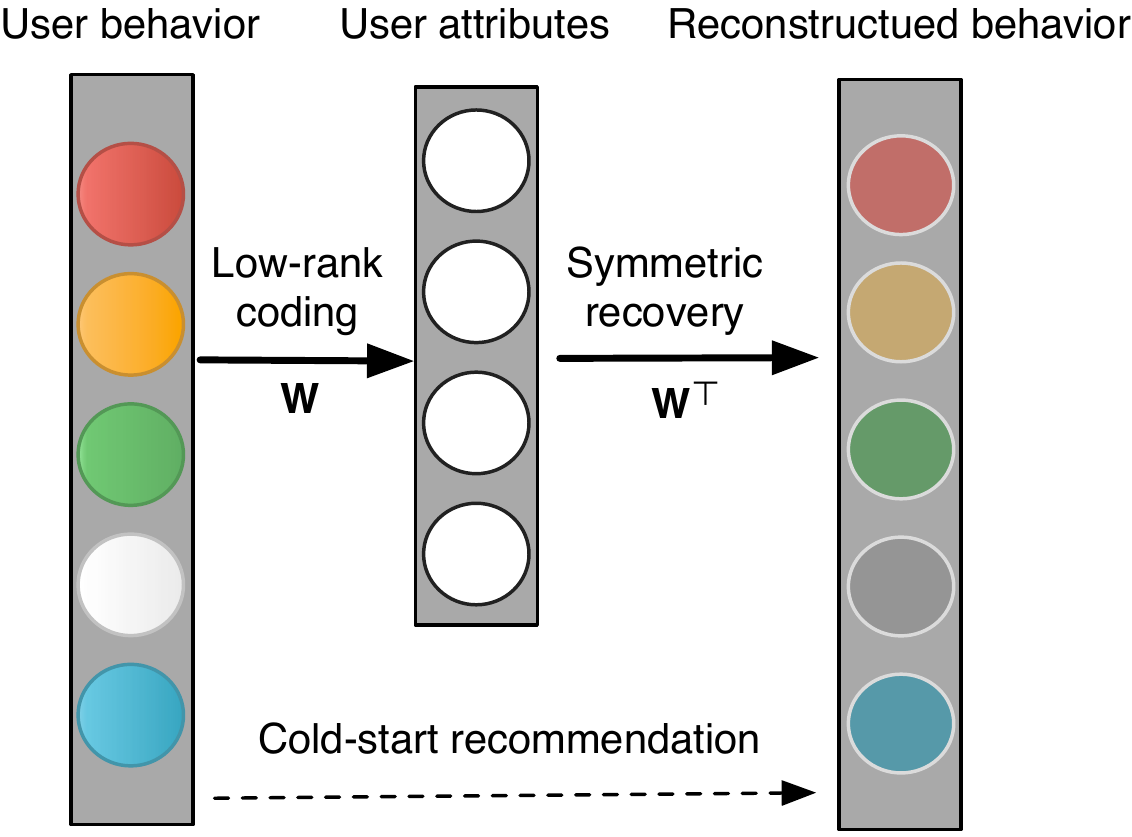}
\end{center}
\vspace{-10pt}
\caption{Illustration of the proposed LLAE. Firstly, we learn a low-rank encoder which maps user behavior into user attributes. Then, user attributes of new users are used to reconstruct the user behavior (generate recommendations for new users). For the sake of efficiency, the parameters of the encoder and decoder are symmetric (tied weights). Notice that the encoder guarantees that warm users and cold users can be compared in the attribute space. The reconstruction, at the same time, assures that the user behavior (recommendation list) can be generated from user attributes.} 
\label{fig:idea}
\end{figure}

Generally, the latent space is represented as hidden layers. For the concern of efficiency and interpretability, we only deploy one hidden layer $\bf S$ in our model. In this paper, $\bf S$ has definite meanings, semantic space in ZSL or user side information in CSR. Recently, tied weights~\cite{mohamed2012acoustic} has been introduced into autoencoders to learn faster models yet with less parameters. In this paper, we consider the tied weights $\bf M=W^\top$. Then, we have the following formulation as illustrated in Fig.~\ref{fig:idea}:

\begin{equation}
\label{eq:llae2}
  \begin{array}{c}
  \min\limits_\mathbf{W} \mathbf{\|X-W^\top WX\|}_F^2, ~~s.t.~\mathbf{WX=S}.
  \end{array} 
\end{equation}

As we stated before, one of the challenge problems in real-world recommender system is that we need to handle very high-dimensional and sparse matrix, because there are millions of items and users but a specific user only have few interactions with few items. To avoid spurious correlations caused by the mapping matrix, we propose that $\bf W$ should be low-rank. As a result, we have:
\begin{equation}
\label{eq:llae3}
  \begin{array}{c}
  \min\limits_\mathbf{W} \mathbf{\|X-W^\top WX\|}_F^2+\beta \mathrm{rank}(\mathbf{W}) , ~~s.t.~\mathbf{WX=S},
  \end{array} 
\end{equation}
where $\mathrm{rank}(\cdot)$ is the rank operator of a matrix, $\beta>0$ is a penalty parameter. It is worth noting that the rank constraint on $\bf W$ benefits the model from at least two aspects. On one side, it helps filter out the spurious correlations from behavior space to attribute space. On the other side, it helps highlight the shared attributes across different users. For instance, a specific attribute, e.g., basketball fan, would be shared by many users from different ages. The low-rank constraint on $\bf W$ will reveal these common attributes.

In some cold-start tasks, the two spaces might be not very correlated. The low-rank constraint helps reveal the most correlated (relatively) part (dominate factors), and the reconstruction part is even more critical because the projection can be spurious in this situation without reconstruction constraint. The reconstruction part is effective in mitigating the domain shift problem. This is because although the user behavior may change from warm users to cold users, the demand for more truthful reconstruction from the attributes to behavior is generalizable across warm and cold domains, resulting in the learned project function less susceptible to domain shift.

The low-rank constraint on $\bf W$ makes the optimization more difficult for the reason that low-rank is a well-known NP-hard problem. As an alternative method, the trace norm $\|\cdot\|_*$ is widely used to encourage low-rankness in previous work~\cite{li2016low}. However, the trace norm controls the single values of the matrix, but the changes of the single values do not always lead to a change of the rank. Inspired by recent work~\cite{dinglow2017}, we propose to use an explicit form of low-rank constraint as follows:
\begin{equation}
\label{eq:llae4}
  \begin{array}{l}
  \min\limits_\mathbf{W} \mathbf{\|X-W^\top WX\|}_F^2+\beta \sum\limits_{i=r+1}^{d} (\sigma_i{(\mathbf{W})})^2 , \\~s.t.~\mathbf{WX=S},
  \end{array} 
\end{equation}
where $\sigma_i{(\mathbf{W})}$ is the $i-$th singular value of $\bf W$, $d$ is the total number of singular values of $\bf W$. Different from the trace norm, $\sum_{i=r+1}^{d} (\sigma_i{(\mathbf{W})})^2$ explicitly solves the problem of minimizing the square sum of $r$-smallest singular value of the mapping $\bf W$.

Note that 
\begin{equation}
  \begin{array}{c}
  \sum_{i=r+1}^{d} (\sigma_i{(\bf W)})^2=\mathrm{tr}\mathbf{(V^\top WW^\top V)},
  \end{array} 
  \label{eq:updateV}
\end{equation}
where $\mathrm{tr}(\cdot)$ is the trace operator of a matrix, and $\bf{V}$ consists of the singular vectors which correspond to the $(d-r)$-smallest singular values of ${\bf WW^\top}$. Thus, our objective function can be written as:
\begin{equation}
\label{eq:llae5}
  \begin{array}{l}
  \min\limits_\mathbf{W,V} \mathbf{\|X-W^\top WX\|}_F^2+\beta \mathrm{tr}\mathbf{(V^\top WW^\top V)} , \\~s.t.~\mathbf{WX=S}.
  \end{array} 
\end{equation}

At last, to learn more robust hidden layers, we train a denoising autoendocoder~\cite{Vincent2008Extracting} by introducing corruptions into the input. Specifically, we randomly set 10\% of ${\bf X}$ to zeros, and get the corrupted version $\widehat{\bf X}$. As a result, we have the final objective:
\begin{equation}
\label{eq:llae6}
  \begin{array}{l}
  \min\limits_\mathbf{W,V} \mathbf{\|X-W^\top WX\|}_F^2+\beta \mathrm{tr}\mathbf{(V^\top WW^\top V)} , \\~s.t.~\mathbf{W \widehat{X}=S}.
  \end{array} 
\end{equation}

\subsection{Problem Optimization}

To optimize Eq.~\eqref{eq:llae6}, we first rewrite it to the following equivalent form:
\begin{equation}
\label{eq:op1}
  \begin{array}{l}
  \min\limits_\mathbf{W,V} \mathbf{\|X-W^\top S\|}_F^2+\beta \mathrm{tr}\mathbf{(V^\top WW^\top V)} , ~s.t.~\mathbf{W \widehat{X}=S}.
  \end{array} 
\end{equation}

However, the constraint on Eq.~\eqref{eq:op1} is hard to optimize. Here we relax the constraint and get the following optimization problem:
\begin{equation}
\label{eq:op2}
  \begin{array}{l}
  \min\limits_\mathbf{W,V} \mathbf{\|X\!-\!W^\top S\|}_F^2\!+\!\lambda\mathbf{\|W \widehat{X}\!-\!S\|}_F^2\!+\!\beta \mathrm{tr}\mathbf{(V^\top WW^\top V)} ,
  \end{array} 
\end{equation}
where $\lambda>0$ is a penalty parameter. As a result, Eq.~\eqref{eq:op2} is a convex problem which has global optimal solution. Then, we calculate the derivative of Eq.~\eqref{eq:op2} with respect to $\bf W$ and set it to zero,
\begin{equation}
\label{eq:op3}
  \begin{array}{l}
  \mathbf{-S(X^\top -S^\top W)+\lambda(WX-S) \widehat{X}^\top+\beta VV^\top W=0},\\ 
  \Rightarrow ~~\mathbf{(SS^\top+\beta VV^\top)W+\lambda WX \widehat{X}^\top=S(X^\top +\lambda\widehat{X}^\top)}
  \end{array} 
\end{equation}

It is worth noting that the optimization of $\bf W$ involves $\bf V$. As an optimization trick~\cite{dinglow2017}, we choose to alternatively update them. At first, by treating $\bf V$ as a constant, we calculate the derivative w.r.t $\bf W$ and set it to zero, as shown in Eq.~\eqref{eq:op3}. Then, we update $\bf V$ by Eq.~\eqref{eq:updateV}.

At last, if we use the following substitutions:
\begin{equation}
\begin{array}{c}
 $$
\begin{cases}
\mathbf{A=SS^\top+\beta VV^\top}\\
\mathbf{B=\lambda X \widehat{X}^\top}\\
\mathbf{C=SX^\top+\lambda S \widehat{X}^\top}\\
\end{cases},$$
\end{array}
\label{eq:m0}
\end{equation}
Eq.~\eqref{eq:op3} can be written as the following equation:
\begin{equation}
\label{eq:op4}
  \begin{array}{l}
  \mathbf{AW+WB=C,}
  \end{array} 
\end{equation}
which can be effectively solved by Sylvester\footnote{www.mathworks.com/help/matlab/ref/sylvester.html} operation in Matlab with only one line of code,
 $$ \mathbf{W=\mathrm{sylvester}(A,B,C)}.$$ For clarity, we show the proposed method in Algorithm~1.

\begin{table}[t]
\centering
\scriptsize
\begin{tabular*}{\linewidth}{@{\extracolsep{\fill}}l}
\hline
{\bf Algorithm 1.} {\it Low-rank Linear AutoEncoder for CSR}\\
\hline
{\bf Input:} User behavior space $\bf X$, user attribute space $\bf S$, parameters $\lambda$ and $\beta$.\\
{\bf Output:} Recommended items for new users.\\
\hline
{\bf Warm-up:}\\
\quad {\it Repeat}\\
\qquad 1. Solve the eigen-decomposition problem to get ${\bf V}:$\\
\qquad \qquad \qquad ${\bf V}\leftarrow \mathrm{svd}({\bf WW}^\top).$\\
\qquad 2. Optimize the encoder $\bf W$ (and the decoder ${\bf W}^\top$):\\
\qquad \qquad \qquad $\mathbf{W=\mathrm{sylvester}(A,B,C)}$, \\
\qquad \quad where $\mathbf{A=SS^\top+\beta VV^\top},~\mathbf{B=\lambda X \widehat{X}^\top},~\mathbf{C=SX^\top+S \widehat{X}^\top}$.\\
\quad {\it Until Convergence}\\

{\bf Cold-start:}\\
\quad  ${\bf X}_{new}={\bf W}^\top {\bf S}_{new}$.\\
{\bf Recommendation:}\\
\quad Using the logistic regression function to predict the recommendation \\\quad probability of items, and recommend the top-$k$ items.\\
\hline
\end{tabular*}
\label{alg:alm}
\end{table}

\subsection{Complexity Analysis}
The computational cost of our algorithm consists of two parts: 1) the optimization of $\bf W$ and 2) the updating of $\bf V$. Generally, both of them cost $O(d^3)$.
However, if we directly calculate $\bf VV^\top$ instead of $\bf V$, the cost of 2) can be reduced to $O(r^2d)$ ($r \ll d$ is the rank of $\bf W$)~\cite{dinglow2017}. In any case, the complexity of our algorithm only depends on the dimensionality. It is irrelevant to the number of samples. As a result, it can be applied to large-scale datasets.

\subsection{Zero-shot Classification}
Given a training data $\bf X$ and a semantic representation $\bf S$, we can learn an encoder $\bf W$ and a decoder $\bf W^\top$ by Eq.~\eqref{eq:op4}. For the new test sample set ${\bf X}_{new}$, we can embed it to a semantic space ${\bf S}_{new}={\bf WX}_{new}$. Then, the labels of ${\bf X}_{new}$ can be learned by a classifier which calculates the distances between ${\bf S}_{new}$ and ${\bf S}_{proto}$, where ${\bf S}_{proto}$ is the projected prototypes in the semantic space.
\begin{equation}
\label{eq:zsl1}
  \begin{array}{l}
  f(\mathbf{X}_{new})=\arg\min{dis}(\mathbf{S}_{new},{\bf S}_{proto}),
  \end{array} 
\end{equation}
where $f(\mathbf{X}_{new})$ is a classifier which returns the labels of ${\bf X}_{new}$, $dis()$ is a distance metric. 

\subsection{Cold-start Recommendation}
Suppose that we use $\bf X$ to denote the user behavior, e.g., purchase, browse and share, which in general is a user-item matrix. $\bf S$ is the user attributes, e.g., user preferences, personal information and social network data. We can learn an encoder $\bf W$ and a decoder $\bf W^\top$ by Eq.~\eqref{eq:op4}. Then, for the new users, CSR aims to learn ${\bf X}_{new}$ which indicates the potential user-item relationship, which can be achieved by:
\begin{equation}
\label{eq:csr1}
  \begin{array}{l}
  \mathbf{X}_{new}=\mathbf{W^\top S}_{new}.
  \end{array} 
\end{equation}

At last, the recommendation will be formulated as a multi-label classification problem~\cite{zhang2014review}. Specifically, we can use the logistic regression function to predict the recommendation probability of items, and recommend the top-$k$ items.

\begin{table}[t!p]
\centering
\caption{Statistics of the tested dataset in ZSL experiments.}
\label{tab:statistics}
\small
\begin{tabular}{|c|c|c|c|c|c|c|c|}
\hline
Dataset & aP\&aY & ~AwA~ & ~CUB~ & ~SUN~ \\
\hline
\# Seen classes   & 20 & 40 & 150 & 707  \\
\hline
\# Unseen classes & 12 & 10 & 50 & 10 \\
\hline
\# Samples    & 15,339 & 30,475 & 11,788 & ~14,340~ \\
\hline
\# Attributes  & 64 & 85 & 312 & 102  \\
\hline
\end{tabular}
\end{table}


\section{Experiments}
In this section, we verify the proposed method on both zero-shot recognition and cold-start recommendation tasks. From Eq.~\eqref{eq:op4}, we know that the main part of our method can be implemented by only one line of Matlab code. The complete codes will be released on publication.

\subsection{Zero-shot Learning}
{\bf Datasets.} For zero-shot recognition, four most popular benchmarks are evaluated. For instance, {\bf aPascal-aYahoo (aP\&aY)}~\cite{farhadi2009describing}, {\bf Animal with Attribute (AwA)}~\cite{lampert2014attribute}, {\bf SUN scene attribute dataset (SUN)}~\cite{patterson2012sun} and {\bf Caltech-UCSD Birds-200-2011 (CUB)}~\cite{wah2011caltech}. The statistics of the datasets are reported in Table~\ref{tab:statistics}.

{\bf Settings.} For ZSL, we follow the experimental protocols reported in previous work~\cite{dinglow2017,kodirov2017semantic}, deep convolutional neural networks (CNNs) features extracted by GoogLeNet~\cite{szegedy2015going}, which is the 1024-dimensional activation of the final pooling layer, are used as input. The hyper-parameters $\lambda$ and $\beta$ are tuned by cross-validation using the training data.

{\bf Baselines.} Five state-of-the-art work, e.g., DAP~\cite{lampert2014attribute}, ESZSL~\cite{romera2015embarrassingly}, SSE~\cite{zhang2015zero}, JLSE~\cite{zhang2016zero} and LESD~\cite{dinglow2017}, are selected as competitors.

\begin{table}[t!p]
\centering
\caption{Accuracy (\%) of zero-shot recognition. The best results are marked as bold numbers.}
\label{tab:zsl}
\small
\begin{tabular}{|c|c|c|c|c|c|c|c|}
\hline
Method & ~aP\&aY~ & ~~AwA~~ & ~CUB~ & ~SUN~ & ~Avg.~\\
\hline
DAP    & 38.23 & 60.51 & 39.14 & 71.92 & 52.45 \\
\hline
ESZSL  & 24.37 & 75.31 & 48.75 & 82.12 & 57.64 \\
\hline
SSE    & 46.22 & 76.35 & 30.49 & 82.51 & 58.89 \\
\hline
JLSE   & 50.46 & 80.51 & 42.83 & 83.86 & 64.42 \\
\hline
LESD   & \bf{58.83} & 76.62 & 56.25 & 88.36 & 70.02 \\
\hline
Ours   & 56.16 & {\bf 85.24} & {\bf 61.93} & {\bf 92.07} & {\bf 73.85} \\
\hline
\end{tabular}
\end{table}

{\bf Results and Discussions.} The experimental results of ZSL are reported in Table~\ref{tab:zsl}. From the results, we can see that our approach, LLAE, performs much better than the compared methods. Since the compared methods cover a wide range of models which deploy different techniques for zero-shot learning, and the state-of-the-art method LESD~\cite{dinglow2017} is reported recently, the performance of our method is quite favorable and significant.

Different from conventional ZSL methods, our model is bilateral. We consider not only the unilateral projection from the feature space to the attribute space but also the reconstruction from the attribute space to the feature space. Although our initial purpose of reconstruction is tailored for cold-start recommendation (one cannot generate the recommendation list without reconstruction from the user information in CSR), it benefits zero-shot recognition as well. The results verify our proposition of this work---ZSL and CSR share the similar problem framework, and the interdisciplinary study on them can benefit both of the communities.

\subsection{Cold-start Recommendation}

{\bf Datasets.} For cold-start recommendation, we mainly use social data as side information. The following four datasets, which consist of image, video, blog and music recommendation, are used for evaluation. 
\begin{itemize}
       \item {\bf Flickr}~\cite{tang2012scalable} is a dataset collected from {\it flickr.com\footnote{http://www.flickr.com}}, which is a popular personal photos managing and sharing website. Users in flickr can tag photos and subscribe photos in terms of tags with which he is interested. For instance, a user can subscribe photos with tag ``baseball''. The evaluated dataset consists of 80,513 users, 195 interest groups as the items, and a social network with 5,899,882 links.
       \item {\bf BlogCatalog}~\cite{tang2012scalable} is a dataset collected from {\it blogcatalog.com}\footnote{http://www.blogcatalog.com}, which is a popular blog collaboration system. Any article published by a blogger in blogcatalog can be cataloged into some groups according to the topics, e.g., ``sports'', ``business'' and ``technology''. The tested dataset consists of 10,312 users, 39 topics as items, and a social network with 333,983 links.
       \item {\bf YouTube}~\cite{tang2012scalable} is a dataset collected from {\it youtube.com}\footnote{http://www.youtube.com}, which is a popular video watching and sharing website. Users in YouTube can also subscribe interested topics. The evaluated dataset consists of 1,138,499 users, 47 categories as items, and a social network with 2,990,443 links.
       \item {\bf Hetrec11-LastFM}~\cite{cantador2011second} is a dataset collected from {\it last.fm}\footnote{http://www.last.fm}, which is an online music system. Hetrec11-LastFM contains social networking, tagging, and music artist listening information. The tested dataset consists of 1,892 users, 17,632 artists as items, and 186,479 tag assignments.
\end{itemize}

{\bf Settings.} For the evaluated datasets, we split each of them into two subsets, one includes 10\% of the users as new users (test dataset) for cold-start, and the remainder of 90\% users are collected as training data to learn the encoder and decoder. We deploy cross-validation with grid-search to tune all hyper-parameters on training data. Specifically, we select 80\% users for training and 10\% for validation. The new users are randomly selected, so we build 10 training-test folds and report the average results.

Following previous work~\cite{sedhain2017low}, we deploy the widely used precision@k, recall@k and mean average precision (mAP@100) as the measurements. All the hyper-parameters in the objective are tuned by cross-validation. The following five previous work powered by different techniques are evaluated as baselines: CBF-KNN~\cite{gantner2010learning}, Cos-Cos~\cite{sedhain2014social}, BPR-Map~\cite{gantner2010learning}, CMF~\cite{krohn2012multi} and LoCo~\cite{sedhain2017low}.

\begin{table}[t]
\centering
\caption{CSR results of mAP@100 (\%) on different datasets.}
\label{tab:map}
\small
\begin{tabular}{|c|c|c|c|c|}
\hline
Method & \hspace{1mm} Flickr \hspace{1mm} & BlogCatalog & YouTube & \hspace{0.2mm} LastFM \hspace{0.2mm}  \\
\hline 
CBF-KNN  & 28.05   & 32.71  & 34.21  & 17.12\\
\hline
Cos-Cos  & 31.42   & 41.06  & 46.67  & 12.26\\
\hline
BPR-Map  & 21.59   & 28.22  & 30.35  & 8.85\\     
\hline
CMF     & 21.41   & 27.76  & 28.16  & 8.13\\
\hline
LoCo     & 33.57   & 45.35  & 48.79  & 18.09\\
\hline
{ Ours} &{\bf 39.25}  & {\bf 50.13}  & {\bf 52.45}  & {\bf 23.07}\\ 
\hline
\end{tabular}
\end{table}

\begin{table*}[ht!p]
\centering
\caption{Cold-start recommendation results (\%) on Flickr dataset.}
\label{tab:flickr}
\small
\begin{tabular}{|c|c|c|c|c|c|c|c|c|c|c|c|c|}
\hline
&\multicolumn{2}{c|}{BPR-Map} & \multicolumn{2}{c|}{CMF} & \multicolumn{2}{c|}{CBF-KNN} & \multicolumn{2}{c|}{Cos-Cos} & \multicolumn{2}{c|}{LoCo} & \multicolumn{2}{c|}{Ours} \\
\hline
 @k& Precision & Recall & Precision & Recall & Precision & Recall & Precision & Recall & Precision & Recall & Precision & Recall \\
 \hline
 1  & 16.99 & 13.52 & 15.91 & 12.34 & 20.62 & 15.79 & 22.33 & 17.58 & 28.64 & 22.52 & {\bf 33.61} & {\bf 26.34}\\
 \hline
 5  & 7.29  & 28.12 &  7.55 & 27.24 & 10.10 & 36.69 & 10.34 & 37.56 & 11.99 & 43.47 & {\bf 15.32} & {\bf 48.27}\\
 \hline
 10 & 4.85  & 37.18 &  5.12 & 36.57 &  6.58 & 47.13 &  6.71 & 47.84 &  7.25 & 51.62 & {\bf 11.03} & {\bf 55.74}\\
 \hline
 20 & 3.28  & 50.21 &  3.28 & 46.67 &  4.01 & 57.17 &  4.12 & 58.24 &  4.12 & 58.31 & {\bf  7.67} & {\bf 62.45}\\
\hline
\end{tabular}
\end{table*}

\begin{table*}[t]
\centering
\caption{Cold-start recommendation results (\%) on BlogCatalog dataset.}
\label{tab:BlogCatalog}
\small
\begin{tabular}{|c|c|c|c|c|c|c|c|c|c|c|c|c|}
\hline
&\multicolumn{2}{c|}{BPR-Map} & \multicolumn{2}{c|}{CMF} & \multicolumn{2}{c|}{CBF-KNN} & \multicolumn{2}{c|}{Cos-Cos} & \multicolumn{2}{c|}{LoCo} & \multicolumn{2}{c|}{Ours} \\
\hline
 @k& Precision & Recall & Precision & Recall & Precision & Recall & Precision & Recall & Precision & Recall & Precision & Recall \\
 \hline
 1  & 16.21 & 11.67 & 18.59 & 15.05 & 20.14 & 16.05 & 31.76 & 25.64 & 37.65 & 30.35 & {\bf 43.55} & {\bf 35.82}\\
 \hline
 5  & 10.13 & 38.89 &  8.62 & 33.32 & 11.38 & 43.14 & 13.26 & 49.78 & 14.26 & 53.24 & {\bf 19.26} & {\bf 57.72}\\
 \hline
 10 &  7.87 & 57.63 &  6.56 & 49.15 &  8.41 & 60.47 &  8.86 & 64.75 &  8.90 & 65.08 & {\bf 12.45} & {\bf 69.24}\\
 \hline
 20 &  5.62 & 81.07 &  4.78 & 69.65 &  5.88 & 83.94 &  5.89 & 84.27 &  5.62 & 80.23 & {\bf  7.36} & {\bf 83.19}\\

\hline
\end{tabular}
\end{table*}

\begin{table*}[ht!p]
\centering
\caption{Cold-start recommendation results (\%) on YouTube dataset.}
\label{tab:youtube}
\small
\begin{tabular}{|c|c|c|c|c|c|c|c|c|c|c|c|c|}
\hline
&\multicolumn{2}{c|}{BPR-Map} & \multicolumn{2}{c|}{CMF} & \multicolumn{2}{c|}{CBF-KNN} & \multicolumn{2}{c|}{Cos-Cos} & \multicolumn{2}{c|}{LoCo} & \multicolumn{2}{c|}{Ours} \\
\hline
 @k& Precision & Recall & Precision & Recall & Precision & Recall & Precision & Recall & Precision & Recall & Precision & Recall \\
 \hline
 1  & 18.76 & 19.15 & 21.18 & 19.15 & 23.58 & 18.23 & 33.86 & 29.15 & 40.22 & 34.35 & {\bf 46.37} & {\bf 39.25}\\
 \hline
 5  & 14.16 & 40.29 & 10.23 & 37.32 & 13.58 & 46.67 & 16.57 & 50.21 & 18.55 & 60.14 & {\bf 22.18} & {\bf 65.23}\\
 \hline
 10 &  9.29 & 59.58 &  8.43 & 53.15 & 10.25 & 63.77 & 10.86 & 67.56 & 12.23 & 69.34 & {\bf 15.39} & {\bf 73.25}\\
 \hline
 20 &  6.25 & 83.51 &  5.91 & 73.25 &  6.67 & 84.61 &  6.84 & 86.59 &  8.77 & 88.74 & {\bf 11.32} & {\bf 92.58}\\

\hline
\end{tabular}
\end{table*}

\begin{table*}[ht!p]
\centering
\caption{Cold-start recommendation results (\%) on Hetrec11-LastFM dataset.}
\label{tab:Hetrec11-LastFM}
\small
\begin{tabular}{|c|c|c|c|c|c|c|c|c|c|c|c|c|}
\hline
&\multicolumn{2}{c|}{BPR-Map} & \multicolumn{2}{c|}{CMF} & \multicolumn{2}{c|}{CBF-KNN} & \multicolumn{2}{c|}{Cos-Cos} & \multicolumn{2}{c|}{LoCo} & \multicolumn{2}{c|}{Ours} \\
\hline
@k& Precision & Recall & Precision & Recall & Precision & Recall & Precision & Recall & Precision & Recall & Precision & Recall \\
 \hline
 1  & 32.53 & 0.81 & 37.54 & 0.94 & 53.39 & 1.35 & 36.54 & 0.92 & 58.11 & 1.42 & {\bf 62.95} & {\bf 4.33}\\
 \hline
 5  & 26.83 & 3.07 & 30.71 & 3.54 & 45.08 & 5.14 & 31.97 & 3.87 & 48.09 & 5.51 & {\bf 55.16} & {\bf 10.27}\\
 \hline
 10 & 23.99 & 5.31 & 25.64 & 5.95 & 39.65 & 8.91 & 28.07 & 6.56 & 41.87 & 9.58 & {\bf 46.67} & {\bf 13.19}\\
 \hline
 20 & 20.53 & 9.05 & 20.97 & 9.67 & 32.32 & 14.39& 22.95 & 10.56& 33.98 & 15.42 & {\bf 39.86} & {\bf 18.45}\\
\hline 
\end{tabular}
\end{table*}

{\bf Results and Discussions.} The experimental results on different datasets are reported in Table~\ref{tab:flickr} - Table~\ref{tab:Hetrec11-LastFM}. Table~\ref{tab:map} shows the mAP@100 on the four datasets. Regarding to the experimental results, we have the following discussions:

1) It has been verified by ZSL that our bilateral formulation, i.e., projection from feature space to attribute space and reconstruction from attribute space to feature space, is very effective. In fact, the bilateral formulation is also the main reason that our model performs better in CSR tasks. As we stated in the section of related work, existing CSR methods generally learn a projection from the user behavior space to the user preference space. The learned projection is unilateral. In other words, the formulations of existing work focus on the mapping from the behavior to the preference, but it did not take reconstruction into consideration. If we take a further look at the CSR problem, we can find that the projection from user behavior to the user preference and the reconstruction from user preference to user behavior is equally important. The projection guarantees that warm users and cold users can be compared in the preference space. The reconstruction assures that the user behavior (recommendation list) can be generated from user preference. 

2) Most of the baselines are based on matrix factorization (MF). The MF approaches, however, is highly sensitive and easily disturbed by noises when the observations are sparse. Compared with the recommendation scenarios where there are intensive feedbacks provided by users and only limited items, social network observations are extremely sparse. LoCo and our LLAE deploy low-rank representation to reveal the true data structure from corrupted inputs. Thus, they outperform the baselines in most evaluations.

3) Although LoCo deploys low-rank representation, it learns a unilateral projection. In social data facilitated CSR problems, there are a large number of users and only limited attributes. Thus, a user will be linked with several attributes, and a specific attribute will be attached with a lot of users. If we learn only one projection, trivial connections will be involved. For example, if we learn only one projection from user behavior to user attributes, a behavior can be linked with almost all of the attributes with a low weight since most of the elements in the projection matrix are not zero. Such a projection will weaken the true attributes, introduce over-fitting and relax the generalization ability. The reconstruction constraint in LLAE can automatically handle this because a bad projection will cause a bad reconstruction.

\begin{figure}[t!p]
\begin{center}
\subfigure{
 \includegraphics[width=0.3\linewidth]{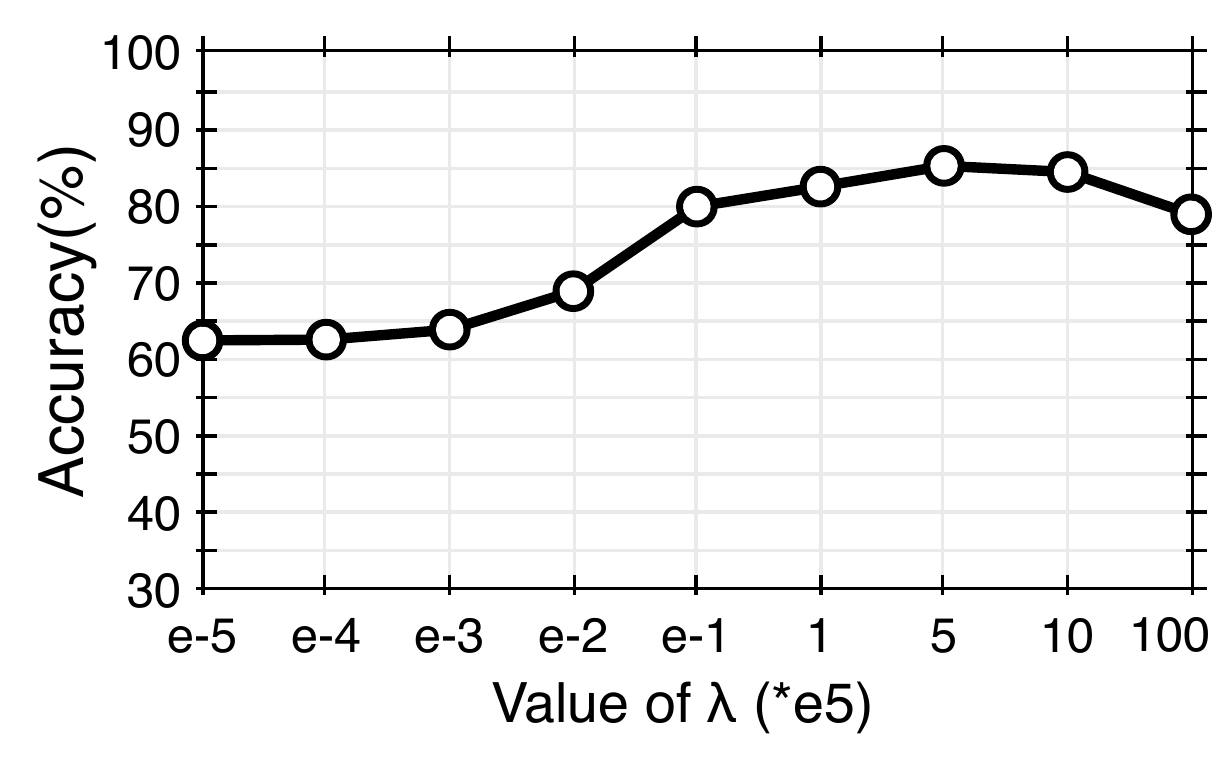}
}
\subfigure{
  \includegraphics[width=0.3\linewidth]{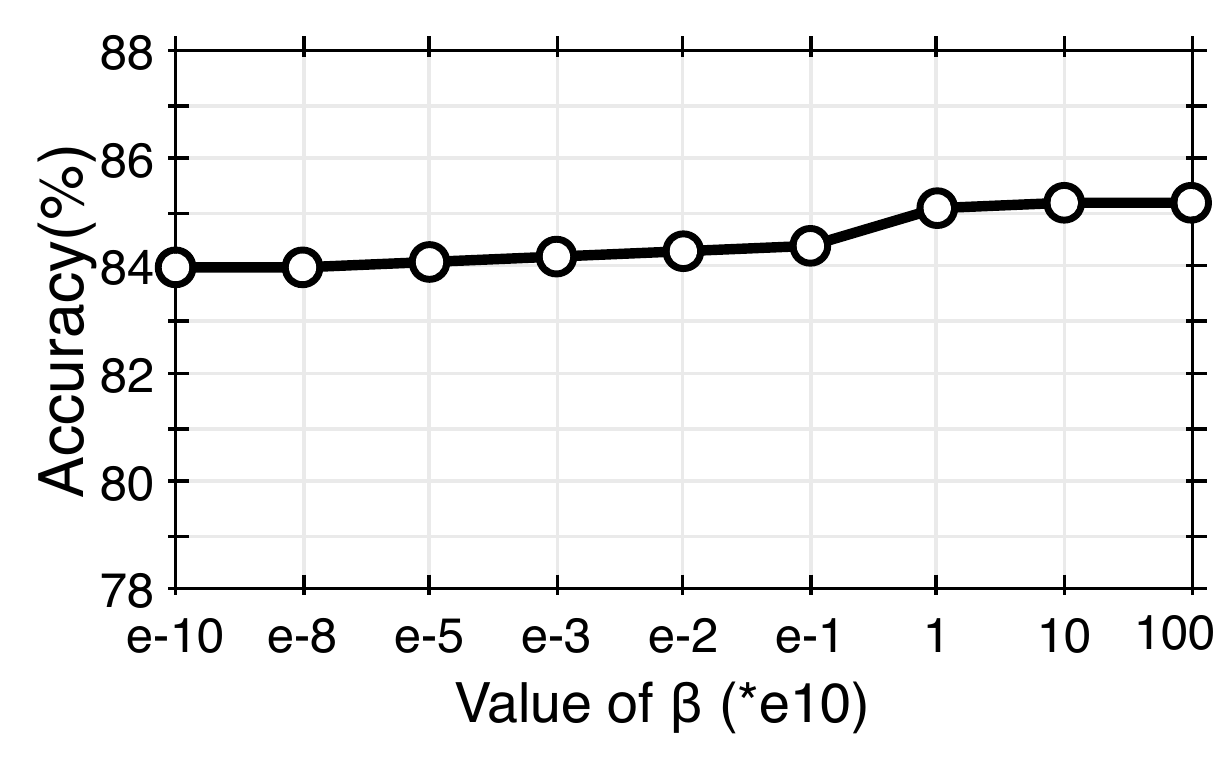}
}
\subfigure{
 \includegraphics[width=0.3\linewidth]{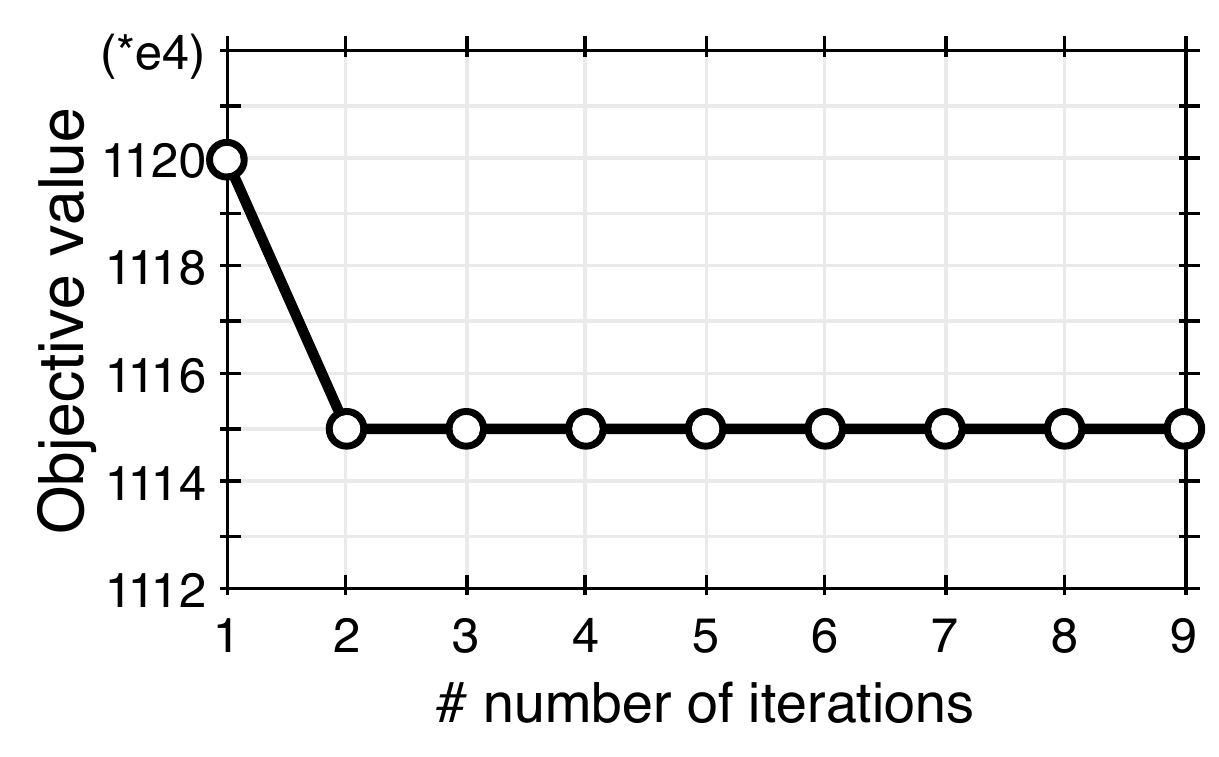}
}
\vspace{-10pt}
\caption{Parameters sensitivity (a-b) and convergence curve (c) of the proposed method on AwA dataset.}
\label{fig:para}
\end{center}
\end{figure}

\begin{figure}[t!p]
\begin{center}
\subfigure[Low-rank]{
 \includegraphics[width=0.46\linewidth]{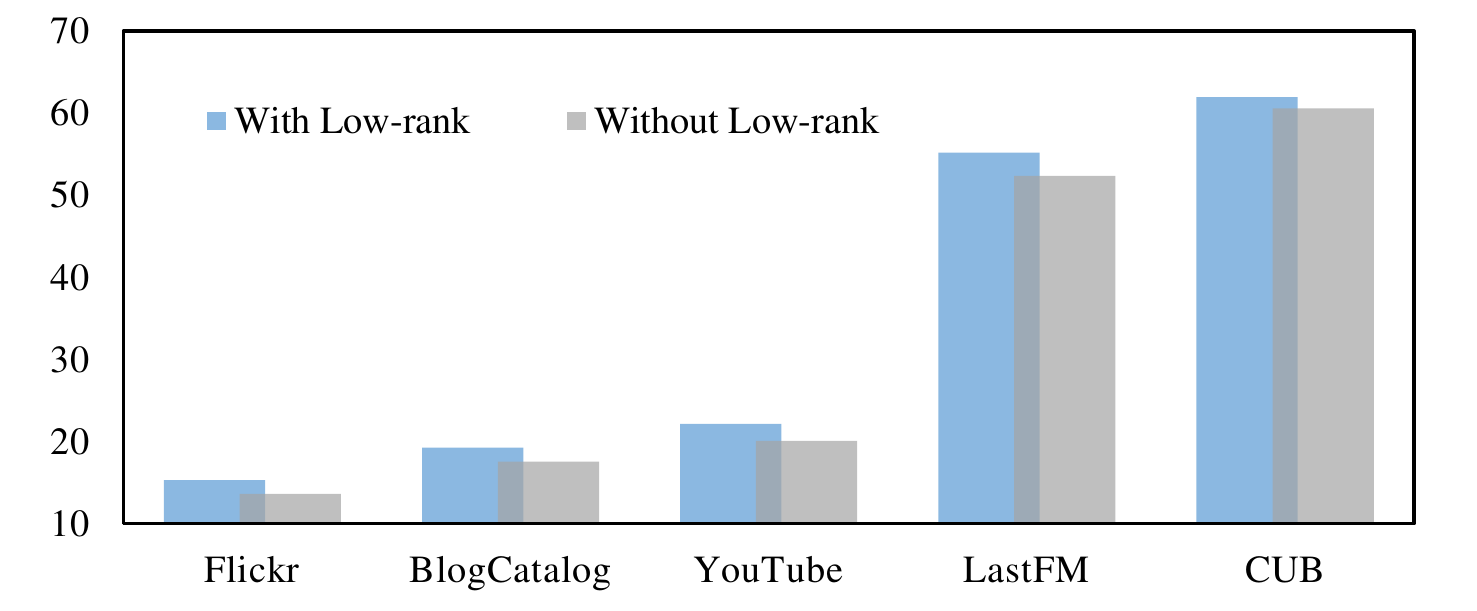}
 }
 \subfigure[Reconstruction]{
 \includegraphics[width=0.46\linewidth]{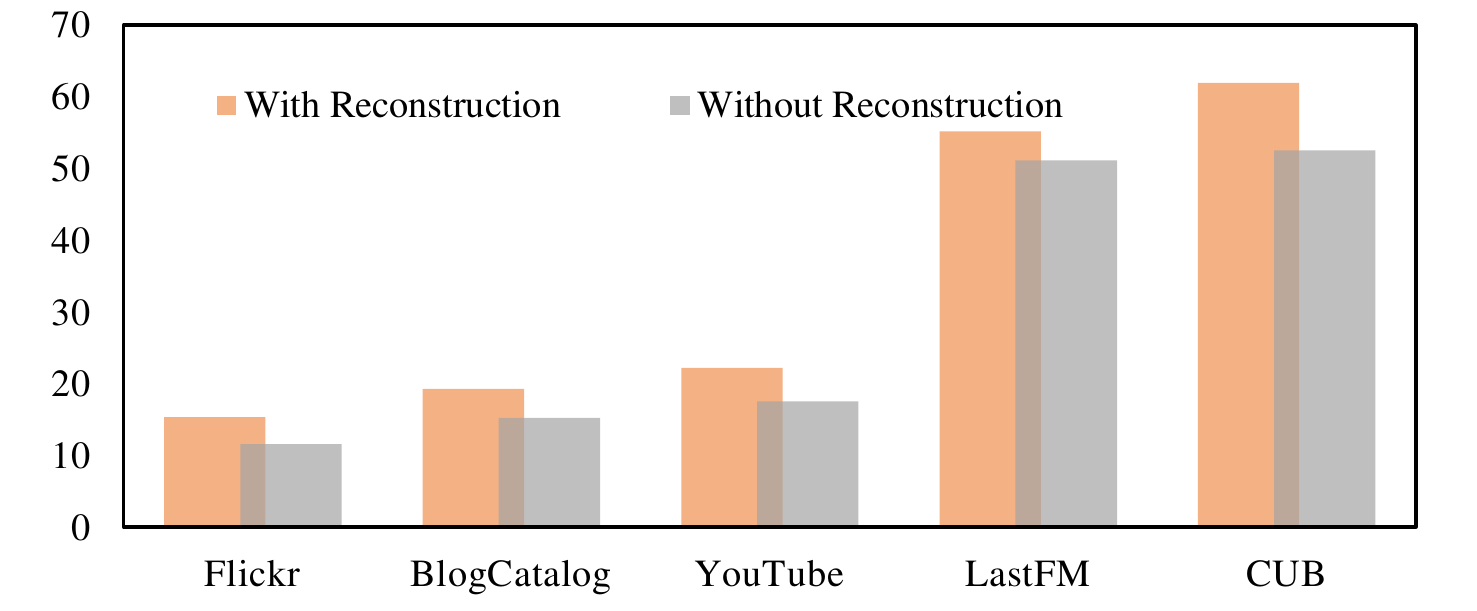}
 }
 \vspace{-10pt}
\caption{The effect of the low-rank and reconstruction constraint. Results of the CSR evaluations are the Precision@5 on different datasets. CUB dataset is used for ZSL.}
\label{fig:low-rank}
\end{center}
\end{figure}

\subsection{Model Analysis}
 {\bf Parameters Sensitivity.} For different dataset, the hyper parameters vary from dataset to dataset. For instance, the optimal value of $\lambda$ on some datasets is around $10^5$, while on others is less than 10. They, therefore, need to be selected by cross-validation. However, when $\lambda$ is large, we also need to increase the value of $\beta$ so that the effect of low-rank constraint would not be neglected (it is easy to be understood by referring to Eq.~\eqref{eq:op4}). Since it is hard to illustrate the effects of the parameters on different datasets (the optimal values can differ by orders of magnitude), the effect of $\lambda$ and $\beta$ on AwA, as an example, are reported in Fig.~\ref{fig:para}(a) and Fig.~\ref{fig:para}(b).

 {\bf Complexity and Convergence.} It is worth noting that LLAE is a linear algorithm, the main part of LLAE can be implemented by only one line of Matlab code, and it runs faster than most of previous work. For instance, it only costs about 1/1000 training time of SSE on AwA. Since we update $\bf W$ and $\bf V$ in an iterative fashion, we show the convergence curve of LLAE on AwA dataset in Fig.~\ref{fig:para}(c). It can be seen that our model converges very fast.

{\bf Low-rank Constraint.} From the parameter curve of $\beta$, we can see that the low-rank constraint is effective for the performance of LLAE. For ZSL, it helps to find out shared semantics across different categories. For CSR, it filters out spurious connections and handles the extremely spare observations. If we investigate the CSR as a specific transfer learning problem~\cite{ding2018incomplete,li2018read}, the low-rank constraint can also mitigate the domain shift between the user behavior space and the user attribute space. We show the effects of the low-rank constraints in Fig.~\ref{fig:low-rank}(a).

{\bf Reconstruction Constraint.} We have mentioned in several places that the reconstruction constraint plays an important role in our formulation. To verify the claim, Fig.~\ref{fig:low-rank}(b) shows the effects of the reconstruction part on several evaluations. It is obvious that the reconstruction constraint contributes a lot for the performance.

\section{Conclusion}
This paper, for the first time, investigates CSR as a ZSL problem. Although CSR and ZSL were independently studied by two communities in general, we reveal that the two tasks are two extensions of the same intension. In this paper, a tailor-made ZSL model is proposed to handle CSR. Specifically, we present a low-rank linear autoencoder, which deploys a low-rank encoder to map user behavior space to user attribute space and a symmetric decoder to reconstruct user behavior from the user attributes. Extensive experiments on eight datasets, including both CSR and ZSL, verify not only that the CSR problem can be addressed by ZSL model, but the consideration of CSR, e.g., the reconstruction constraint, can benefit ZSL as well. It is a win-win formulation. At last, by linking CSR and ZSL, we wish that this work will benefit both of the communities and elicit more contributions. In our future work, we are going to investigate training deep autoencoders for both ZSL and CSR. 

\section{Acknowledgments}
This work was supported in part by the National Natural Science Foundation of China under Grant 61806039, 61832001, 61802236, 61572108 and 61632007, in part by the ARC under Grant FT130101530, in part by the National Postdoctoral Program for Innovative Talents under Grant BX201700045, and in part by the China Postdoctoral Science Foundation under Grant 2017M623006.

\bibliography{aaai19}
\bibliographystyle{aaai}

\end{document}